\documentclass{article}
\usepackage{spconf,amsmath,graphicx}
\usepackage{array}
\usepackage{multirow}
\newcolumntype{L}{>{\arraybackslash}m{17mm}}
\newcolumntype{M}{>{\centering\arraybackslash}m{3mm}}
\newcolumntype{R}{>{\centering\arraybackslash}m{4mm}}

\title{Enhanced Object Detection via Fusion With Prior Beliefs from Image Classification}

\name{Yilun Cao\sthanks{The first two authors contributed equally to this paper.}$^{\dagger}$, Hyungtae Lee\footnotemark[1]$^{\ddagger\mathsection}$, and Heesung Kwon$^{\mathsection}$}
\address{$^{\dagger}$University of Southern California, Los Angeles, California, U.S.A.\\
$^{\ddagger}$Booz Allen Hamilton Inc., McLean, Virginia, U.S.A.\\
$^{\mathsection}$U.S. Army Research Laboratory, Adelphi, Maryland, U.S.A.\\
{\small\tt yiluncao@usc.edu, lee\_hyungtae@bah.com, heesung.kwon.civ@mail.mil}}

\begin{document}
\maketitle
\begin{abstract}
In this paper, we introduce a novel fusion method that can enhance object detection performance by fusing decisions from two different types of computer vision tasks: object detection and image classification. 
In the proposed work, the class label of an image obtained from image classification is viewed as prior knowledge about existence or non-existence of certain objects.  The prior knowledge is then fused with the decisions of object detection to improve detection accuracy by mitigating false positives of an object detector that are strongly contradicted with the prior knowledge.
A recently introduced novel fusion approach called dynamic belief fusion (DBF) is used to fuse the detector output with the classification prior.  Experimental results show that the detection performance of all the detection algorithms used in the proposed work is improved on benchmark datasets via the proposed fusion framework. 


\end{abstract}

\section{Introduction}

Object detection is a fundamental problem in computer vision where one must localize and identify objects of interest in an image.  Over the past decade, many algorithms have been developed to solve this problem such as support vector machine with histograms of oriented gradients (HOG-SVM) \cite{hogsvm}, deformable part models (DPM) \cite{dpm}, deep convolutional neural networks (DCNN)-based detectors \cite{rcnn,fastrcnn,fasterrcnn,resnet,deeploc}.  
In addition to efforts in algorithm development, attempts have been made to improve performance through preprocessing (e.g. object proposals \cite{rcnn,dpm_cascade}), post-processing (e.g. bounding box refinement \cite{bbox_refine,bbox_refine_icip}), and fusing the output of several algorithms (i.e. late fusion \cite{dbf}).
In particular, using late fusion approaches can be quite advantageous when the selected detection algorithms are complementary to each other, resulting in improved fusion performance.  In the proposed work, a relatively unconventional approach compared to previous fusion methods is used that combines the outputs of two different types of computer vision tasks: object detection and image classification.  

\begin{figure}[t]

\begin{minipage}[b]{1.0\linewidth}
  \centering
  \centerline{\includegraphics[width=\textwidth,trim=5mm 15mm 5mm 15mm]{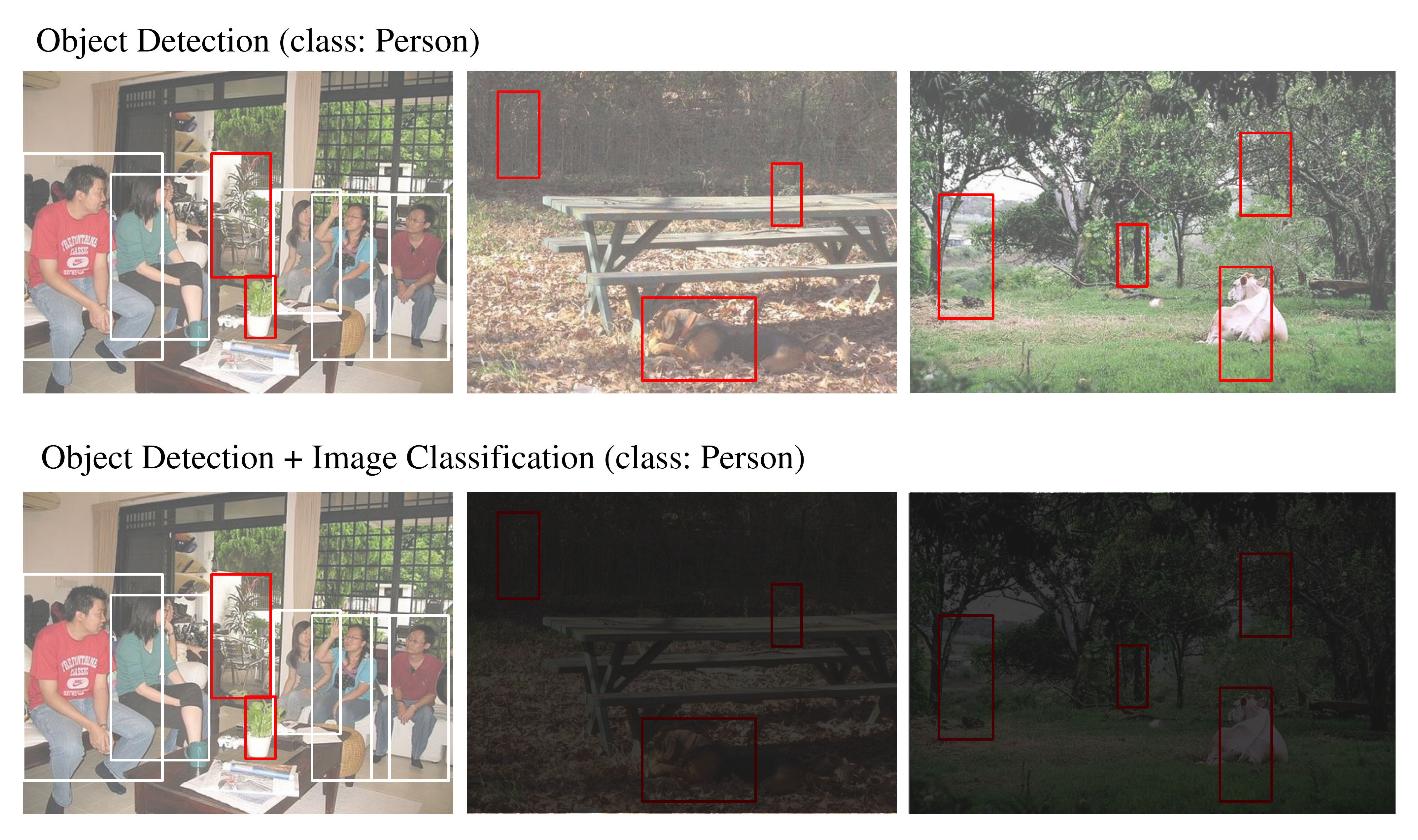}}
\end{minipage}

\caption{{\bf The proposed fusion concept:} the first row presents detection results from a person detector.  In the second row, only the left-most image is determined as a `person' image after fusing detection and classification results.  True positives and false positives are indicated with white and red bounding boxes, respectively.  Note that the precision values of the first and second row are 5/14 and 5/7, respectively.  Precision can be increased by fusing object detection and  image classification.}
\label{fig:intro}
\end{figure}

\begin{figure*}[t]

\begin{minipage}[b]{1.0\linewidth}
  \centering
  \centerline{\includegraphics[width=0.9\textwidth,trim=0mm 25mm 0mm 15mm]{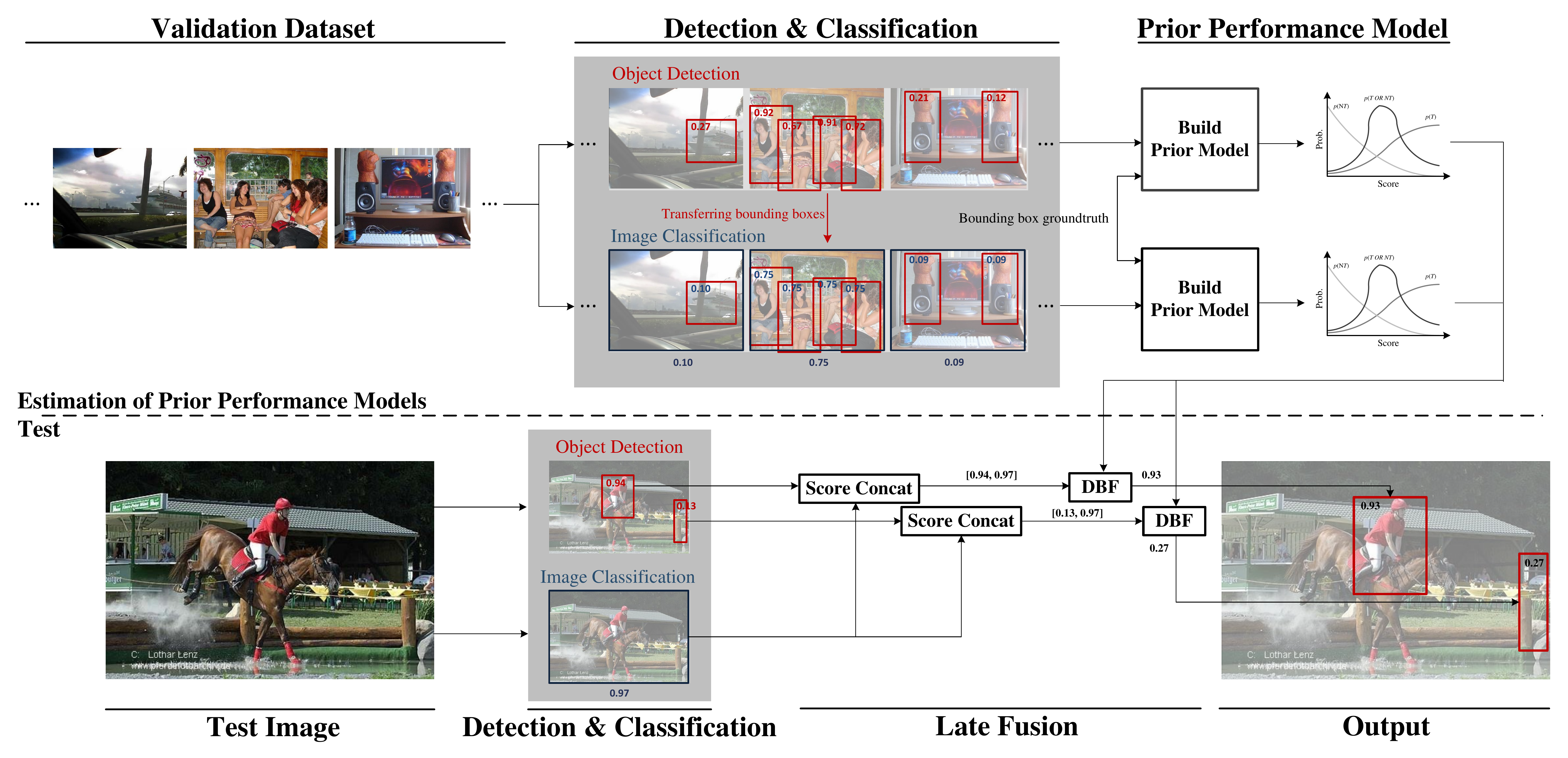}}
\end{minipage}
\caption{Illustration of the overall process of the proposed fusion framework.}
\label{fig:overall}
\end{figure*}

Image classification aims to determine the label of an image by calculating the likelihood of the image to include objects of certain classes.  Even though image classification methods cannot localize objects of interest in an image, they can still provide useful information in the form of a degree of confidence about whether the image contains certain objects or not.  The degree of confidence from classifiers can basically be viewed as prior knowledge about the existence (or non-existence) of the object classes in the image.  This prior knowledge becomes very valuable when fused with the outputs of object detectors as it can possibly remove some false positives of object detectors if the prior knowledge strongly contradicts them.  Similarly, the prior knowledge reinforces the findings of true positives if it strongly agrees with current detection results.  Therefore, fusing the outputs of object detectors and classifiers can possibly enhance precision, a ratio between the number of true positive and the number of true positives and false positives together, eventually improving average precision (AP).  Figure~\ref{fig:intro} presents our idea to use image classification to enhance object detection performance.  

Optimally integrating decisions from object detectors and image classifiers is also a key to enhancing detection performance. Lee et al.~\cite{dbf} recently introduced a late fusion approach, called dynamic belief fusion (DBF), which can effectively integrate decisions from multiple complementary object detection algorithms providing enhanced fusion performance. 
DBF basically assigns probabilities to detection-relevant hypotheses, which are {\it target}, {\it non-target}, and {\it target OR non-target}, based on confidence levels in the detection results conditioned on the prior performance of individual algorithms.  For object detection, DBF clusters bounding boxes from multiple algorithms and outputs a fusion score for the cluster. However, since a image classification approach cannot localize objects in an image, a strategy is needed that can convert classification scores to detection scores associated with bounding boxes. We use a relatively simple strategy that assigns a classification score of each image equally to all the bounding boxes (object candidates) found by object detectors from the same image.  For object detection, three detection algorithms with varying degrees of performance are selected: HOG-SVM \cite{hogsvm}, DPM \cite{dpm}, and Faster R-CNN \cite{fasterrcnn}. We also use a weakly supervised convolutional neural network (WCNN) \cite{wcnn} as an image classification algorithm. 

Our contributions are summarized as follows:
\begin{enumerate}
\item We introduce a novel fusion framework that can enhance detection performance of object detectors by using prior knowledge about existence or non-existence of certain objects in an image estimated from image classification.
\item To the best of our knowledge, the proposed fusion approach is the first attempt to combine detection and classification tasks to improve detection accuracy of current state-of-the-art detection approaches.
\end{enumerate}

\section{The Proposed Approach}

\subsection{Overview}

The proposed fusion framework consists of three steps: 
(i) training an object detection algorithm and an image classification algorithm, 
(ii) estimating prior performance models for individual algorithms, and 
(iii) integrating the outputs of individual algorithms by using DBF, a novel fusion algorithm previously developed by two of the authors.
The dataset is divided into three non-overlapping subsets ({\it train} / {\it validation} / {\it test}).  Note that both the object detection algorithm and the image classification algorithm are trained on {\it train} dataset, and {\it validation} and {\it test} sets are used for prior performance modeling and performance evaluation, respectively.  The overall fusion process of the proposed work is illustrated in Figure~\ref{fig:overall}.\\[0.05cm]

\noindent \textbf{Estimation of Prior Performance Models:} With DBF, the prior performance models of an individual object detection algorithm and an image classification algorithm are estimated from the validation set.  The prior models are estimated in the form of the precision-recall (PR) relationship to represent a level of prior confidence of both the detection and the classification algorithms.  To calculate the PR curve for the object detector, all detection windows are labelled as {\it true} or {\it false positives} in reference to ground truth bounding boxes.  If the intersection-over-overlap between a detection window and the corresponding ground truth bounding box is over a certain threshold (e.g. 0.5), the detection is labeled as {\it true positive}, otherwise {\it false positive}.  To calculate equivalent object detection performance of an image classifier, the output score of the classifier is converted to a detection score by assigning the classification score to all the detection windows equally found by a object detector for the same image.\\[0.05cm]

\noindent \textbf{Test:} For each detection window of an object detector, the corresponding detection scores from both the detector and the classifier are concatenated to form a score vector, which is used as an input to DBF.  DBF then estimates a fused score of the corresponding detection by integrating the score vector, a current observation, with the prior confidence models of both the detection and the classification algorithms.

\subsection{Dynamic Belief Fusion (DBF)}

To effectively fuse detection scores of each detection window from individual detector and classifier, a novel fusion method proposed by Lee et al.~\cite{dbf} called Dynamic Belief Fusion (DBF) is used to build a probabilistic fusion model. 

\begin{figure}[t]

\begin{minipage}[b]{1.0\linewidth}
  \centering
  \centerline{\includegraphics[width=0.9\textwidth,trim=0mm 15mm 0mm 15mm]{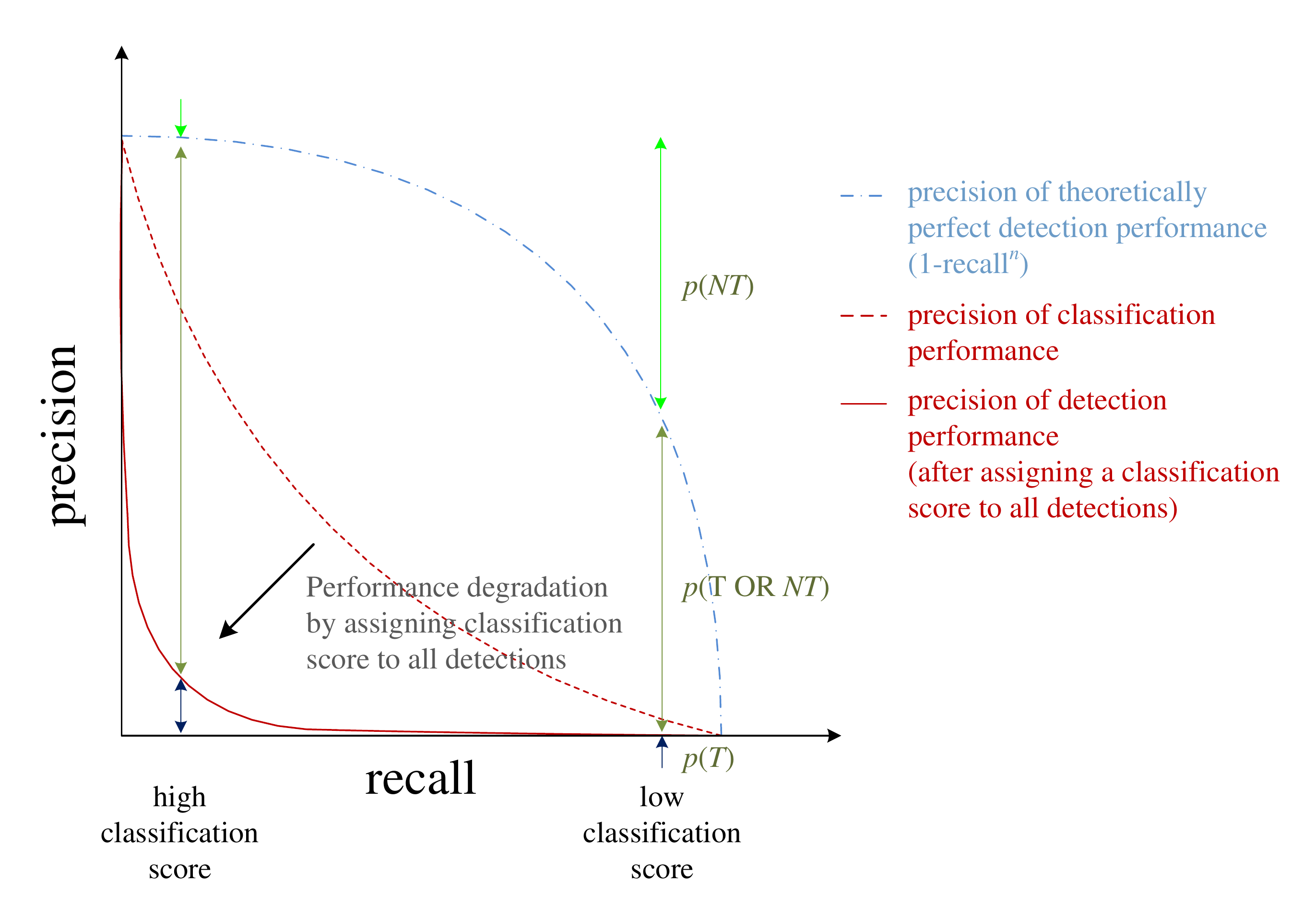}}
\end{minipage}
\caption{The prior performance model of an individual algorithm. The plot also shows the probability assignments for {\it target}, {\it target or non-target}, and {\it non-target} hypotheses for an image classification algorithm.  The equivalent detection precision of the classification algorithm is estimated by assigning the classification score to all the detections of a detector.}
\label{fig:back_sup}

\end{figure}

For a two-class object detection problem, DBF uses a set of hypotheses defined as $\{T,~NT,~T~OR~NT\}$, where $T$ and $NT$ are a {\it target} and {\it non-target} hypothesis, respectively.  $T~OR~NT$ represents detection ambiguity, which indicates that the subject observation could be either {\it target} or {\it non-target}.  For each detection, the corresponding probabilities are assigned to the three hypotheses by linking the current detection score to the prior performance model of each detector, as shown in Figure \ref{fig:back_sup}.  The prior performance model is basically a precision-recall relationship estimated from the validation dataset.  Then the probabilities assigned to the three hypotheses are defined as
\begin{eqnarray}
p(T) &=& prec(s) \\\nonumber
p(NT) &=& rec^n(s) \\\nonumber
p(T~OR~NT) &=& 1-prec(s)-rec^n(s),
\end{eqnarray}
, where $prec$ and $rec$ are precision and recall of the prior performance model, respectively.  $s$ is a detection score. 

Once, for each algorithm, each detection score is converted to the probabilities of the three hypotheses, a final fused probability is calculated by using Dempster's combination rule, which is defined as
\begin{equation}
p_1\oplus p_2(c|s_1,s_2) = \frac{1}{L}\sum_{X\cap Y=c, c\neq\emptyset}{p_1(X|s_1)p_2(Y|s_2)},
\end{equation}
where $L=\sum_{X \cap Y \neq\emptyset} p_1(X|s_1) p_2(Y|s_2)$ is a normalization term. $X,~Y,~c$ can be any hypothesis from a set of $\{T,~NT,~T~OR~NT\}$.  $p(T)-p(NT)$ becomes the fusion out score.  More details are described in \cite{dbf}.\\[0.05cm]

\noindent{\bf Effect of the Reduction of False Positives in the Background Image:} Figure~\ref{fig:back_sup} shows that for a classification algorithm, the degradation of detection precision caused by false positives occurs mainly by assigning a classification score directly to all detection windows.  Note that the precision degradation lowers $p(T)$ but does not affect $p(NT)$ since $p(NT)$ is only defined by recall that does not depend on false positives.  For a test image with a high classification score does not affect detection accuracy because of relatively low $p(T)$ and $p(NT)$.  But, for a test image with a relatively low classification score, a strong prior of $NT$, a large value of $p(NT)$ is assigned, which suppresses all the detections from a detector as false positives.


\section{Experiments}

\subsection{Experimental Setup}

\noindent \textbf{Datasets:} The proposed work is evaluated on the PASCAL VOC 2007 \cite{voc2007} and VOC 2012 \cite{voc2012}, which have been widely used for evaluating object detection performance.  VOC 2007 contains $\sim$ 2.5k images in {\tt train} set, $\sim$ 2.5k images in {\tt val} set, $\sim$ 5k images in {\tt test} set, and $\sim$ 25k object annotations. VOC 2012 is a similar dataset with approximately twice the number of the images and objects in VOC 2007.  \cite{dbf} use a dataset partition of {\tt train}/{\tt val}/{\tt test} to avoid overfitting in building prior performance models.  In addition to this partition, we also evaluate with a common partition of {\tt trainval}/{\tt trainval}/{\tt test} to validate output of individual algorithms by comparing to the performance reported in the original literatures of selected detection algorithms. \\[0.05cm]

%

%

\noindent \textbf{Image Classification:} We use a recently introduced weakly supervised convolutional neural network (WCNN) \cite{wcnn}, which provides significantly enhanced performance in image classification on the PASCAL VOC dataset.  The architecture of WCNN consists of 9 convolutional layers, the first five of which are pre-trained on ImageNet \cite{imagenet}.  All 9 layers are further fine-tuned to the PASCAL VOC 2007 and 2012 datasets.  
In WCNN, an input image is first decomposed into multi-scale images to which multi-scale CNN networks are applied.  The classification scores from individual multiscale CNN pipelines are averaged together to produce a final output score.  Since training the CNN pipelines does not require bounding box labels, the algorithm is called weakly supervised CNN.\\[0.05cm]

\noindent \textbf{Object Detection:} For object detection algorithms, three different algorithms with varying degrees of performance are used: (i) support vector machine with histograms of oriented gradient features (HOG-SVM) \cite{hogsvm}, (ii) deformable part models (DPM) \cite{dpm}, which represents objects as a collection of local parts, and (iii) faster R-CNN \cite{fasterrcnn}, which is the current state-of-the-art in object detection and also runs in real time.

\subsection{VOC 2007 and 2012 Results}

\begin{table}[t]

\begin{center}
\caption{{\bf VOC2007 detection performance:} The mean of average precision (mAP) across all object categories is used as an evaluation metric.  FUSION indicates fusing a object detector with an image classification, WCNN.  FR RCN is Faster R-CNN with ZF net \cite{zfnet}.}
\label{tab:voc07}
{\small
\begin{tabular}{l|l|l|l|l}
\hline
method & train set & val set & mAP & gain \\\hline\hline
HOG-SVM & {\tt train} &  & .184 & \\
FUSION & {\tt train} & {\tt val} & .248 & + .064 \\\hline
DPM & {\tt train} &  & .222 & \\
FUSION & {\tt train} & {\tt val} & .237 & + .015 \\\hline
FR RCN & {\tt train} &  & .585 & \\
FUSION & {\tt train} & {\tt val} & .607 & + .022 \\\hline\hline
HOG-SVM & {\tt trainval} &  & .228 & \\
FUSION & {\tt trainval} & {\tt trainval} & .303 & + .075 \\\hline
DPM & {\tt trainval} &  & .312 & \\
FUSION & {\tt trainval} & {\tt trainval} & .343 & + .031\\\hline
FR RCN & {\tt trainval} &  & .643 & \\
FUSION & {\tt trainval} & {\tt trainval} & .660 & + .017\\\hline
\end{tabular}
}
\end{center}
\end{table}

\begin{table}[t]

\begin{center}
\caption{{\bf VOC2012 detection performance:} The mean of average precision (mAP) across all object categories is used as an evaluation metric.  FUSION indicates fusing a object detector with an image classification, WCNN.  FR RCN is Faster R-CNN with ZF net \cite{zfnet}.}
\label{tab:voc12}
{\small
\begin{tabular}{l|l|l|l|l}
\hline
method & train set & val set & mAP & gain \\\hline\hline
HOG-SVM & {\tt train} &  & .179 & \\
FUSION & {\tt train} & {\tt val} & .250 & + .071 \\\hline
DPM & {\tt train} &  & .258 & \\
FUSION & {\tt train} & {\tt val} & .312 & + .054 \\\hline
FR RCN & {\tt train} &  & .534 & \\
FUSION & {\tt train} & {\tt val} & .553 & + .019 \\\hline\hline
HOG-SVM & {\tt trainval} &  & .203 & \\
FUSION & {\tt trainval} & {\tt trainval} & .277 & + .074 \\\hline
DPM & {\tt trainval} &  & .288 & \\
FUSION & {\tt trainval} & {\tt trainval} & .349 & + .061 \\\hline
FR RCN & {\tt trainval} &  & .569 & \\
FUSION & {\tt trainval} & {\tt trainval} & .583 & + .014 \\\hline
\end{tabular}
}
\end{center}
\end{table}

\noindent \textbf{Detection Accuracy:} Tables~\ref{tab:voc07} and \ref{tab:voc12} show the detection performance of the three detection algorithms as well as the fusion performance with WCNN on both VOC 2007 and VOC 2012, respectively.  It is shown that all the detection algorithms benefit from fusing with WCNN, via DBF on both datasets.  The fusion gain is mainly attributed to the reduction of false positives, objects of non-interest recognized as objects of interest by the detectors.  The false positives that are strongly contradicted with the  classification prior indicating a high level of likelihood of non-existence of certain objects are basically eliminated through the fusion process.  It is observed that the fusion gain for HOG-SVM is much greater than DPM and FR-RCN.  This is because HOG-SVM results in more false positives than DPM and FR-RCN, many of which are removed using the prior knowledge from WCNN.\\[0.05cm]

\noindent \textbf{Dataset Partitions:} We use two different dataset partitions for evaluation.  The first partition, which is {\tt train} / {\tt val} / {\tt test}, avoids overfitting while optimizing both training detectors/classifier and building prior performance models.  The second partition, which is {\tt trainval} / {\tt trainval} / {\tt test} allows the overfitting.  However, the evidence of performance degradation by overfitting has not been observed.

\section{Conclusions}

We have introduced a novel fusion framework that can enhance the detection accuracy of existing object detection algorithms by fusing with the prior knowledge about the existence (or non-existence) of certain objects obtained from image classification. In the proposed work, we mainly focus on mitigating false positives from object detection by using the proposed fusion strategy that can eliminate any false positive strongly contradicted with the prior knowledge estimated from image classification. The experimental results in Tables \ref{tab:voc07} and \ref{tab:voc12} show that the reduction of false positives via the proposed fusion approach directly leads to enhanced detection accuracy. It is also observed that the proposed fusion with the detection algorithms with more false positives, such as HOG-SVM and DPM, provides greater fusion gain than the fusion with faster R-CNN. This confirms the basic premise of the proposed fusion strategy that the prior knowledge from image classification can effectively reduce false positives.


\bibliographystyle{IEEEbib}
\bibliography{references}

\begin{thebibliography}{10}

\bibitem{hogsvm}
Navneet Dalal and Bill Triggs,
\newblock ``Histograms of oriented gradients for human detection,''
\newblock in {\em IEEE Conference on Computer Vision and Pattern Recognition},
  2005.

\bibitem{dpm}
Pedro~F. Felzenszwalb, Ross~B. Girshick, David McAllester, and Deva Ramanan,
\newblock ``Object detection with discriminatively trained part based models,''
\newblock {\em IEEE Transactions on Pattern Analysis and Machine Intelligence},
  vol. 32, no. 9, pp. 1627--1645, 2010.

\bibitem{rcnn}
R.~B. Girshick, J.~Donahue, T.~Darrell, and J.~Malik,
\newblock ``Rich feature hierarchies for accurate object detection and semantic
  segmentation,''
\newblock in {\em IEEE Conference on Computer Vision and Pattern Recognition},
  2014.

\bibitem{fastrcnn}
Ross Girshick,
\newblock ``Fast r-cnn,''
\newblock in {\em IEEE International Conference on Computer Vision}, 2015.

\bibitem{fasterrcnn}
Shaoqing Ren, Kaiming He, Ross~B. Girshick, and Jian Sun,
\newblock ``Faster {R-CNN}: Towards real-time object detection with region
  proposal networks,''
\newblock in {\em Advances in Neural Information Processing Systems}, 2015.

\bibitem{resnet}
Kaiming He, Xiangyu Zhang, Shaoqing Ren, and Jian Sun,
\newblock ``Deep residual learning for image recognition,''
\newblock in {\em IEEE Conference on Computer Vision and Pattern Recognition},
  2015.

\bibitem{deeploc}
Archith~J. Bency, Heesung Kwon, Hyungtae Lee, S~Karthikeyan, and B.~S.
  Manjunath,
\newblock ``Weakly supervised localization using deep feature maps,''
\newblock in {\em European Conference on Computer Vision}, 2016.

\bibitem{dpm_cascade}
Pedro~F. Felzenszwalb, Ross~B. Girshick, and David McAllester,
\newblock ``Cascade object detection with deformable part models,''
\newblock in {\em IEEE Conference on Computer Vision and Pattern Recognition},
  2010.

\bibitem{bbox_refine}
Spyros Gidaris and Nikos Komodakis,
\newblock ``Object detection via a multi-region \& semantic segmentation-aware
  cnn model,''
\newblock in {\em IEEE International Conference on Computer Vision}, 2015.

\bibitem{bbox_refine_icip}
Kai-Wen Cheng, Yie-Tarng Chen, and Wen-Hsien Fang,
\newblock ``Iterative locallization refinement in convolutional neural networks
  for improved object detection,''
\newblock in {\em IEEE International Conference on Image Processing}, 2016.

\bibitem{dbf}
Hyungtae Lee, Heesung Kwon, Ryan~M. Robinson, William~D. Nothwang, and Amar~M.
  Marathe,
\newblock ``Dynamic belief fusion for object detection,''
\newblock in {\em IEEE Winter Conference on Applications of Computer Vision},
  2016.

\bibitem{wcnn}
Maxime Oquab, L\'eon Bottou, Ivan Laptev, and Josef Sivic,
\newblock ``Is object localization for free? – weakly-supervised learning
  with convolutional neural networks,''
\newblock in {\em IEEE Conference on Computer Vision and Pattern Recognition},
  2015.

\bibitem{voc2007}
Mark Everingham, Luc Van~Gool, Christopher K.~I. Williams, John Winn, and
  Andrew Zisserman,
\newblock ``The {PASCAL} {V}isual {O}bject {C}lasses {C}hallenge 2007
  {(VOC2007)} {R}esults,''
  http://www.pascal-network.org/challenges/VOC/voc2007/workshop/index.html.

\bibitem{voc2012}
Mark Everingham, S.~M.~Ali Eslami, Luc Van~Gool, Christopher K.~I. Williams,
  John Winn, and Andrew Zisserman,
\newblock ``The pascal visual object classes challenge: A retrospective,''
\newblock {\em International Journal of Computer Vision}, vol. 111, no. 1, pp.
  98--136, Jan. 2015.

\bibitem{imagenet}
Olga Russakovsky, Jia Deng, Hao Su, Jonathan Krause, Sanjeev Satheesh, Sean Ma,
  Zhiheng Huang, Andrej Karpathy, A.~Khosla, Michael Bernstein, Alexander~C.
  Berg, and Li~Fei-Fei,
\newblock ``{ImageNet Large Scale Visual Recognition Challenge},''
\newblock {\em International Journal of Computer Vision}, vol. 115, no. 3, pp.
  211--252, 2015.

\bibitem{zfnet}
Matthew~D. Zeiler and Rob Fergus,
\newblock ``Visualizing and understanding convolutional networks,''
\newblock in {\em European Conference on Computer Vision}, 2014, pp. 818--833.

\end{thebibliography}

\end{document}